\documentclass[journal]{IEEEtran}
\usepackage{graphics} % for pdf, bitmapped graphics files
\usepackage{epsfig} % for postscript graphics files
\usepackage{mathptmx} % assumes new font selection scheme installed
\usepackage{amsmath} % assumes amsmath package installed
\usepackage{amssymb}  % assumes amsmath package installed
\usepackage[utf8]{inputenc}
\usepackage[T1]{fontenc}
\usepackage{mathtools}
\usepackage{enumitem}
\usepackage{xcolor}
\usepackage{float}
\usepackage{gensymb}
\usepackage{bm}
\usepackage{autobreak}
\usepackage{svg}
\usepackage{mathrsfs}
\usepackage{tabularx}
\usepackage{array, makecell}
\usepackage{graphicx}
\usepackage{cellspace}
\usepackage{textcomp}
\usepackage{hyperref}
\usepackage{cleveref}

\usepackage{comment}

\newcommand{\cusst}[1]{{}}

\DeclareMathAlphabet{\mathcal}{OMS}{cmsy}{m}{n}  % thus can use mathcal{}

\begin{document}

\title{
Concentric Tube Robot Redundancy Resolution via Velocity/Compliance Manipulability Optimization
}

\author{Jia~Shen*, Yifan Wang*, Milad Azizkhani, Deqiang  Qiu, and Yue Chen% <-this % stops a space
\thanks{This research is supported by Georgia Tech faculty startup grant and McCamish Blue Sky Grant. Corresponding author: Yue Chen. }
\thanks{J. Shen, Y. Wang, and M. Azizkhani are with the Department of Mechanical Engineering, Georgia Institute of Technology, Atlanta 30332 USA (e-mail: \{jshen359, wangyf, mazizkhani3\}@gatech.edu.)}% <-this % stops a space
\thanks{D. Qiu is with the Department of Radiology and Imaging Sciences, Emory University, Atlanta 30338 USA (e-mail: deqiang.qiu@emory.edu)}

\thanks{Y. Chen is with the Department of Biomedical Engineering, Georgia Institute of Technology/Emory, Atlanta 30332 USA (e-mail: yue.chen@bme.gatech.edu)}

\thanks{* Jia Shen and  Yifan Wang contributed equally to this paper.} 
\thanks{This work has been submitted to the IEEE for possible publication. Copyright may be transferred without notice, after which this version may no longer be accessible.}

% <-this % stops a space
}


\markboth{}%
{Shell \MakeLowercase{\textit{et al.}}: Bare Demo of IEEEtran.cls for IEEE Journals}



\maketitle

\begin{abstract}
Concentric Tube Robots (CTR) have the potential to enable effective minimally invasive surgeries. While extensive modeling and control schemes have been proposed in the past decade, limited efforts have been made to improve the trajectory tracking performance from the perspective of manipulability
, which can be critical to generate safe motion and feasible actuator commands. In this paper, we propose a gradient-based redundancy resolution framework that optimizes velocity/compliance manipulability-based performance indices during trajectory tracking
for a kinematically redundant CTR. 
We efficiently calculate the gradients of manipulabilities by propagating the first- and second-order derivatives of state variables of the Cosserat rod model
along the CTR arc length, reducing the gradient computation time by 68\% compared to finite difference method. Task-specific performance indices are optimized by projecting the gradient into the null-space of trajectory tracking. 
The proposed method is validated in three exemplary scenarios that involve trajectory tracking, obstacle avoidance, and external load compensation, respectively.
Simulation results show that the proposed method is able to accomplish the required tasks while commonly used redundancy resolution approaches underperform or even fail.

\end{abstract}

% Note that keywords are not normally used for peer review papers.
\begin{IEEEkeywords}
Concentric Tube Robot,  Manipulability, Redundancy Resolution
\end{IEEEkeywords}

% \fi
\IEEEpeerreviewmaketitle

\section{Introduction}

\IEEEPARstart{C}{oncentric} Tube Robots (CTR) consist of concentrically aligned, pre-curved elastic tubes, and  
are capable of generating dexterous motions. The dexterity and compact dimension of these devices make them ideal for a variety of minimally invasive surgical applications \cite{mitros2022review}.
Extensive research has been conducted on the mechanics modeling of CTR, aiming to characterize the mapping from the joint-space input to  robot configuration. The most widely adopted approach combines the Cosserat rod model with geometric concentric constraints of tubes
\cite{rucker2010kinematics, Dupont_2010_CTR_model}. This approach describes the spatial evolution of robot states with a system of ordinary differential equations (ODEs), resulting in a boundary value problem (BVP) that can be numerically solved by nonlinear root-finding algorithms. The Cosserat-based model has been used to formulate and solve problems of stability analysis \cite{gilbert2015elastic}, stiffness modulation \cite{xiao2023kinematics},  and force sensing \cite{xiao2023curvature}.

Despite significant advancements over the past decade, achieving reliable trajectory tracking with CTR still presents a significant technical challenge, primarily due to the difficulties in accurately solving the complicated inverse kinematics \cite{mahoney2019review}. 
This statement remains particularly true when CTRs are required to operate in complex scenarios, such as being in close proximity to singular configurations or when secondary tasks like obstacle avoidance and carrying external loads are necessary.
The Jacobian-based resolved rate controller and many of its variations have been widely adopted to partially address this problem. 
Recent progress includes the efficient Jacobian calculation via forward integration approach \cite{rucker2011computing}, and redundancy resolution for secondary task optimization such as joint limit avoidance \cite{burgner2013telerobotic}, and instability avoidance \cite{AndersonRedundancyICRA17}. 
The damped least squares approach \cite{DLS_singularity_1986} can be used to prevent undesirable behavior of the robot under ill-conditioned Jacobians, but the additional regulation term may drive the robot away from the desired trajectory, leading to unwanted behavior.
% \ma{this sentence is not accurate, please modify}. 
Note that the external force disturbances may cause significant deflections of the CTR in unfavorable configurations, which can contribute to inaccurate trajectory tracking. 
 
The velocity/compliance manipulability  is a crucial performance measure to evaluate the robot singularity and force capacity for a given configuration \cite{CTR_manipulability}, which is essential for the safe and efficient manipulation of CTR in confined environments or having contacts. The concept of manipulability was originally proposed for rigid-link robots \cite{yoshikawa1985manipulability} to determine whether the posture is compatible with task requirements. Despite the significant structural differences between CTR and rigid-link robots, the concept of manipulability can be generalized to CTR \cite{CTR_velo_manipulability}. A unified force/velocity manipulability index for CTR was proposed in \cite{CTR_manipulability} to estimate the optimal direction for a better force/velocity transmission ratio. However, there have been limited efforts to address the trajectory tracking problem by considering CTR  manipulability. 
%\yf{Hessian-based kinematic control review \cite{gradients_calc}}
One of the most recent studies used the gradient projection method, which tries to reshape the unified compliance/velocity manipulability ellipsoid into a sphere along the trajectory to avoid instability  \cite{khadem2019RR_manipul}. However, this method only considers the velocity manipulability at the tip for instability avoidance, and the Hessian is approximated using the Broyden-Fletcher-Goldfarb-Shanno (BFGS) method, which introduces errors that may cause unstable performance and slow down the optimization process \cite{bfgs_nocedal1999}.

In this paper, we present a gradient-based redundancy resolution framework for CTR that optimizes motion/force capability along any task-required direction during trajectory tracking. We develop the derivative propagation method for the gradient of manipulability, enabling efficient calculation and online trajectory tracking as needed. 
Furthermore, we propose several task-specific performance indices based on velocity/compliance manipulability, which are optimized by gradient projection. The performance of the redundancy resolution framework is demonstrated through simulations of a three-tube CTR, where the robot is controlled to follow a desired trajectory while utilizing the redundant degrees of freedom (DoFs) to accomplish secondary tasks, including singularity avoidance, obstacle avoidance, and stiffness modulation.
This paper is organized as follows: Section \ref{preliminaries} provides an overview of the CTR  model and manipulability indices. The derivative propagation method is presented in Section \ref{Hessian}. Section \ref{section_controller} details the redundancy resolution with task-specific performance indices. The simulation results are presented in Section \ref{simulation}, followed by the conclusion in Section \ref{conclusion}.

\section{Preliminaries} \label{preliminaries}
\subsection{Review of CTR Mechanics Model} \label{CTR_model} 
The use of Cosserat rod theory for modeling the mechanics of CTR is a widely accepted approach \cite{rucker2010kinematics}. This section provides a brief overview of the CTR mechanics model, and Table 1 summarizes the nomenclature used in this paper. We refer the reader to \cite{rucker2011computing} for detailed derivation.

\begin{figure}[t!]
    \centering
    \includegraphics[width = 1.00\linewidth]{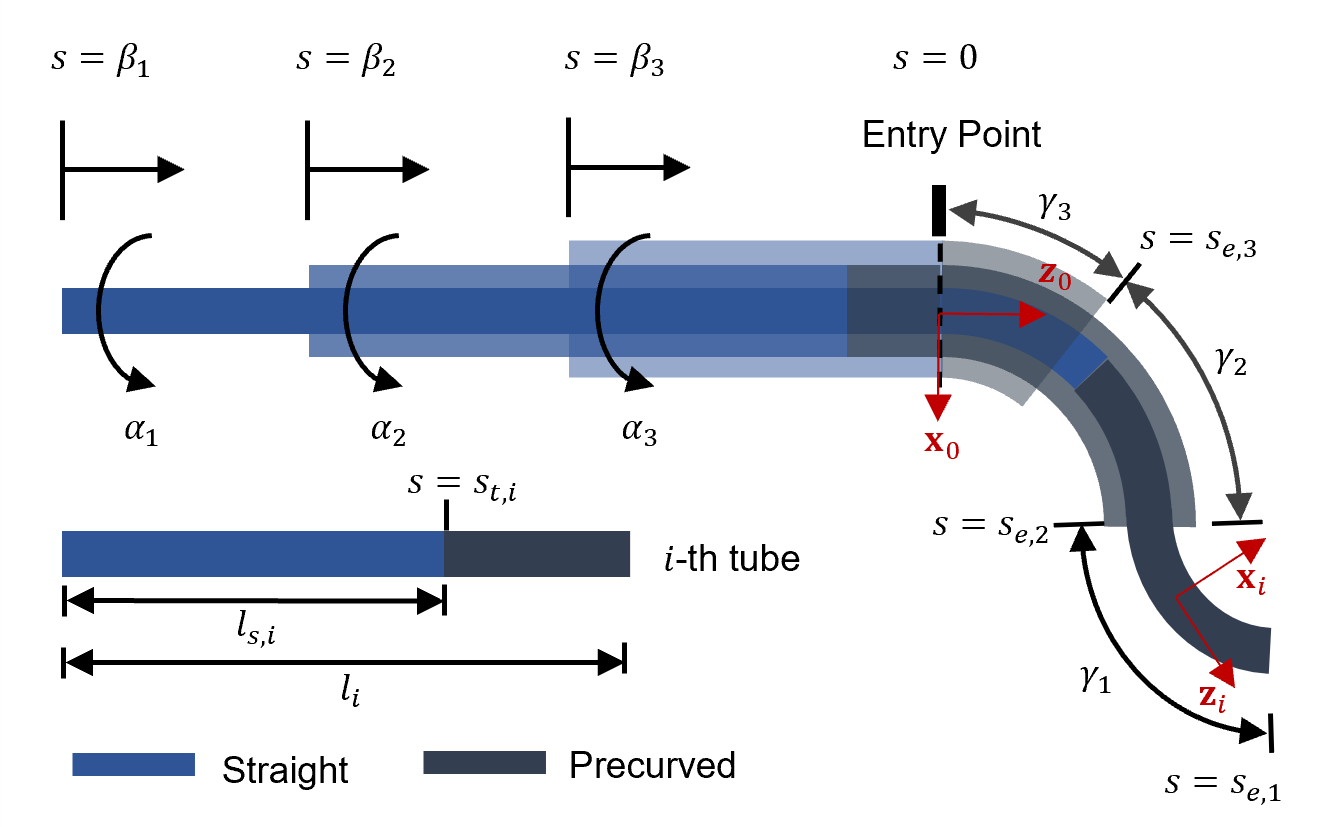}
    \caption{A CTR with three precurved tubes.  The base rotation, translation, and exposed length of $i$-th tube are denoted as $\alpha_i$, $\beta_i$, and $\gamma_i$,  respectively. } 
    \label{fig:CTR_illustration}
    \vspace{- 2mm}
\end{figure}

As shown in Fig.\ref{fig:CTR_illustration}, the shape of a CTR is described as a differentiable spatial curve parameterized by its arc length $s$. A material frame is assigned to each tube such that the origin of the frame moves along the curve at $s$, and the $z$-axis of the frame aligns with the tangent of the curve. Assuming that all tubes conform to the same curve, the position of all material frames w.r.t. the fixed reference frame is given by $\mathbf{p}(s)$ and the orientation of the $i$-th tube is given by $\mathbf{R}_i(s)$. To simplify notation, we use $\mathbf{R}_1(s) = \mathbf{R}(s)$. It then follows that $\mathbf{R}_i(s) = \mathbf{R}(s) \mathbf{R}_z(\theta_i(s))$, where $\mathbf{R}_z(\theta)$ denotes the rotation around $z$-axis for angle $\theta$. The curvature of the $i$-th tube at $s$ represents the rate of change of $\mathbf{R}_i(s)$ w.r.t. $s$, $\mathbf{u}_i(s) = \left ( \mathbf{R}_i(s)^T \mathbf{R}_i^{\prime}(s) \right )^{\vee}$.

\begin{table}[th!]
\renewcommand{\arraystretch}{1.2}
\caption{Nomenclature}
\label{notation}
\centering
\begin{tabularx}{8.7cm}{|l|X|}
\hline 
 \textbf{Notation} & \textbf{Definitions}\\
\hline
$i$ & Tube index, the innermost tube is $i=1$.\\
 $s$ &  Arc-length parameter for the central axis \\
 $l_i$ &  Length of the $i$-th tube \\
$l_{s,i}$  & Length of the straight part of the $i$-th tube \\
 $\alpha_i$  & Rotation angle of the $i$-th tube \\
 $\beta_i$  & Translation length of the $i$-th tube \\
 $s_{e,i}$  & Arc-length parameter at the end of the $i$-th tube: \\
& $s_{e,i} = \beta_i + l_i$ \\
 $s_{t,i}$  & Arc-length parameter at the transition from the straight to the precurved part of the $i$-th tube: $s_{t,i} = \beta_i + l_{s,i}$ \\
$\gamma_i$  & Exposed length of the $i$-th tube, $\gamma_i = s_{e,i} - s_{e,i+1}$ \\%\yf{show this in fig.1} %\yf{conflicts with vector x} 
 $\mathbf{q}$  & $\mathbf{q} = \left [\alpha_1~ \beta_1~ ...~ \alpha_n~ \beta_n \right ]^T$ Actuation vector \\
 $\mathbf{p}(s)$  & Position vector of material frames w.r.t. the reference frame \\
 $\mathbf{R}_i(s)$  & {Rotation matrix of the $i$-th material frame w.r.t. the reference frame}\\
 $\mathbf{u}_i(s)$  & {Curvature of the $i$-th tube  w.r.t. the $i$-th material frame}\\
 $\mathbf{m}_i(s)$ & {Internal moment of the $i$-th tube  w.r.t. the reference frame }\\
 $\mathbf{m}^b(s)$  & Total internal moment w.r.t. the first material frame \\
 $\theta_i(s)$  & Rotation angle from the first tube to the $i$-th tube \\ % around the central axis \\
 $\mathbf{F}$ & External force applied to the tip of CTR \\
 $\mathbf{L}$ & External moment applied to the tip of CTR \\
 $\mathbf{e}_3$  & Unit vector of $z$-axis: $[0~0~1]^T$ \\
 $E$ & Young's Modulus \\
 $G$ & Shear Modulus \\
 $I_i$ & {Second moment of area of the cross-section of the $i$-th tube} \\
$J_i$ & { Polar moment of inertia of the cross-section of the $i$-th tube} \\
$c \theta$, $s \theta$  & Concise notations of $\cos \theta$ and $\sin \theta $ \\
$\partial_{x}$, $\partial^2_{x,y}$  & Concise notations of $\dfrac{\partial }{\partial x}$ and $\dfrac{\partial^2 }{\partial x \partial y}$ \\
 $(\cdot)^{\wedge}$  & {Mapping of a vector in $\mathbb{R}^3$ and $\mathbb{R}^6$ to the corresponding element in $\mathfrak{so}(3)$ and $\mathfrak{se}(3)$, respectively.} \\
 $(\cdot)^{\vee}$ &  Inverse operation of $(\cdot)^{\wedge}$ \\
$(\cdot)_{xy}$ &  $\mathbb{R}^3$ to $\mathbb{R}^2$,
extraction of the first two dimension \\ %\yf{$\mathbf{R}^3$ 
% $\mathbf{\mathbf{R}}^3$ vector, \\
$(\cdot)^{\prime}$ &  Derivative w.r.t. arc-length parameter $s$ \\
$\otimes$ & {Tensor product between a tensor and a matrix that contracts the specific dimensions}, e.g.\\
 & if $\mathbf{W} = \mathbf{U} \otimes \mathbf{V}$, then $\displaystyle \mathbf{W}_{i,j,m} = \sum_{k} \mathbf{U}_{i,j,k} \cdot \mathbf{V}_{k,m}$ \\ 
\hline
\end{tabularx}
\vspace{- 6mm}
\end{table}

Consider an $n$-tube CTR with wrench $\mathbf{w} = [\mathbf{F}^T~\mathbf{L}^T]^T$ applied to the tip. The tubes are assumed to have planar precurvature $\mathbf{u}^{*}_i(s) = [\kappa_i,0,0]^T$. It is also assumed that there are no shear and extension in the tubes and no friction between the tubes, which is widely adopted in the literature \cite{rucker2010kinematics}, \cite{Dupont_2010_CTR_model}. Cosserat rod model of CTR describes the evolution of the curve and internal moment by a system of ODEs consisting of geometric constraints, moment equilibrium, and linear constitutive laws as follows:
\begin{subequations} \label{ode_kinematics}
    \begin{align}
        \textbf{p}^{\prime} & = \mathbf{R} \textbf{e}_3 \\
        \mathbf{R}^{\prime} & = \mathbf{R} \hat{\mathbf{u}}_1    \\
        \theta^{\prime}_i & = u_{iz} - u_{1z} \, , \quad i = 2...n \\
        u^{\prime}_{iz} & = - \frac{\kappa_i E I_i}{G J_i} u_{iy} \, , \quad i = 1...n \\
        {\textbf{m}_{xy}^b}^{\prime} & = (-\hat{\textbf{u}}_1 \textbf{m}^b - \hat{\mathbf{e}}_3 \textbf{R}^T \mathbf{F})_{xy} \, ,
    \end{align}
\end{subequations}

\vspace{- 2mm}
The unknown variables on the right-hand side are given by

\vspace{- 6mm}
\begin{subequations} \label{kinematics_relations}
    \begin{align}
        m_z^b & = G \sum_{i=1}^n J_i u_{iz} \\
        \textbf{u}_{1xy} & = \frac{1}{(\sum_{i=1}^n E I_i)} (\textbf{m}^b_{xy} + \sum_{i=1}^n{[c \theta_i \, ~ s \theta_i]^T E I_i \kappa_i}) \\
        u_{iy} & = [-s \theta_i \, ~ c \theta_i] \textbf{u}_{1xy  } 
    \end{align}
\end{subequations}
We can write (\ref{ode_kinematics}) in a compact form:
\begin{subequations} \label{compact_odes}
    \begin{align}
        \mathbf{g}^\prime(s) &= \begin{bmatrix}
        \mathbf{R}(s) & \mathbf{p}(s) \\
        \mathbf{0}_{1\times3} & 1 
        \end{bmatrix}^\prime = \mathbf{g}\Hat{\mathbf{\zeta}}(\mathbf{y}) \\
        \mathbf{y}^{\prime} & = \mathbf{f}(s, \mathbf{y},\mathbf{R},\mathbf{w}) 
    \end{align}
\end{subequations}
where $\mathbf{\zeta} = [\mathbf{v}^T~\mathbf{u}^T]^T$ is the body twist of the material frame w.r.t. $s$, and $\textbf{y} = [\theta_2~ ...~ \theta_n~ u_{1z}~ ...~ u_{nz}~ m_x^b~ m_y^b]^T$. The above ODE  is constrained  at the robot base and the end of each tube. The initial conditions at the robot base are given by geometric constraints determined by $\mathbf{q}$ and $u_{iz}$:
\begin{subequations} \label{initial_cond}
    \begin{align}
        \mathbf{p}(0) & = [0, 0, 0]^T \, \\
        \mathbf{R}(0) & = \mathbf{R}_z(\alpha_1 - \beta_1 u_{1z}(0)) \, \\
        \theta_i(0) & = \alpha_i - \alpha_1  - (\beta_i u_{iz}(0) - \beta_1 u_{1z}(0)) 
    \end{align}
\end{subequations}
The boundary constraints come from the moment equilibrium at the end of each tube, which can be summarized into a vector form:
\begin{equation} \label{boundary_definition}
    \begin{aligned}
        \mathbf{0} = \mathbf{b}(\mathbf{x}) & = [
         G J_1 u_{1z}(s_{e,1}) - \mathbf{e}_3^T \textbf{R}^T \mathbf{L} \, , \\
         & \quad u_{2z}(s_{e,2}) , ..., u_{nz}(s_{e,n}) \, , \\
         & \quad \mathbf{m}_{xy}^b(s_{e,1}) - \left (\mathbf{R}^T \mathbf{L} \right ) _{xy} ]^T. \\
    \end{aligned}
\end{equation}
where the vector $\textbf{x} = [\mathbf{q}^T~ \mathbf{w}^T~\mathbf{y}_u(0)^T]^T$ contains the independent system inputs $\mathbf{q}$ and $\mathbf{w}$, as well as the unknown initial variables $\mathbf{y}_u(0) = [u_{1z}(0)~ ...~ u_{nz}(0)~ m_x^b(0)~ m_y^b(0)]^T$. Equations (\ref{compact_odes})-(\ref{boundary_definition}) form a BVP that can be solved using shooting method, which uses nonlinear root-finding algorithms to iteratively search for the $\mathbf{y}_u(0)$ that satisfy $\mathbf{b}(\mathbf{x}) = \mathbf{0}$. In each iteration, $\mathbf{b}(\mathbf{x})$ is obtained by solving an initial value problem (IVP) consisting  $\mathbf{y}_u(0)$ together with equations (\ref{compact_odes})-(\ref{initial_cond}).

\subsection{Manipulability Analysis}

To characterize the robot versatility of moving in the task space, the notion of velocity manipulability ellipsoid (VME) is proposed in \cite{yoshikawa1985manipulability}. It is defined as 
\begin{equation} \label{VME_def}
\text{VME} := \{\mathbf{\xi}~|~\mathbf{\xi} = \mathbf{J}\Dot{\mathbf{q}}, ||\mathbf{\dot{q}}||=1\}
\end{equation}
where $\mathbf{J} = \left[ \left ( (\text{d}_{q_1} \textbf{g}) \textbf{g}^{-1} \right )^{\vee}~ ...~
    \left ( (\text{d}_{q_n} \textbf{g}) \textbf{g}^{-1} \right )^{\vee} \right ]$ is the spatial Jacobian that maps a unit sphere of joint space velocity to the ellipsoid of task space velocity. 
The velocity manipulability index (VMI) $\mu_v$ is then defined to be the volume of the VME:
\begin{equation} \label{velo_manipul_def}
    \mu_v = \sqrt{\text{det}(\mathbf{J} \mathbf{J}^T)}
\end{equation}

Similarly, the compliance manipulability ellipsoid (CME) is defined as 
\begin{equation} \label{CME_def}
\text{CME} := \{\mathbf{\xi}~|~\mathbf{\xi} =\mathbf{C}\Dot{\mathbf{w}},||\mathbf{\dot{w}}||=1\}
\end{equation}
where $\mathbf{C} = \left[ \left ( (\text{d}_{w_1} \textbf{g}) \textbf{g}^{-1} \right )^{\vee} 
    ~...~ \left ( (\text{d}_{w_6} \textbf{g}) \textbf{g}^{-1} \right )^{\vee} \right ]$ is the compliance matrix of the robot. And the 
 compliance manipulability index (CMI) $\mu_c$ is defined as:
\begin{equation}
    \mu_c = \sqrt{\text{det}(\mathbf{C} \mathbf{C}^T)}
\end{equation}

Note that VMI and CMI are functions of $\mathbf{J}$ and $\mathbf{C}$, respectively. To optimize the manipulability using redundancy resolution, we need to calculate the gradients of $\mathbf{J}$ and $\mathbf{C}$ w.r.t. $\mathbf{q}$, i.e. the Hessians. However, for the Cosserat rod models, a closed-form expression of $\mathbf{J}$ and $\mathbf{C}$ usually are not available.  A feasible way to compute their gradient is using finite difference but it can be computationally expensive. To reduce the heavy computational load, we propose an efficient method for calculating the Hessian below.

\section{Derivative Propagation for the Hessian} \label{Hessian}

 Our calculation of the Hessian adopts the idea of derivative propagation, which essentially combines the propagation of system state variables together with their derivatives into a new system of ODEs. In \cite{rucker2011computing}, an augmented IVP was defined which, in addition to (\ref{ode_kinematics}), includes the propagation of first-order derivatives of state variables along the arc length, to efficiently compute the Jacobian of the CTR. We extend this derivative propagation technique to second-order derivatives, allowing the calculation of the Hessian by solving a single IVP after solving the BVP for $\mathbf{y}_u(0)$, which facilitates the manipulability optimization for redundancy resolution.

We first find the formulation for the Jacobian and compliance matrices. The changes in actuation variables $\mathbf{q}$ and external wrench $\mathbf{w}$ contribute to the spatial twist $\xi = \left ( \mathbf{\dot{g}} \mathbf{g}^{-1} \right )^{\vee}$, where $\mathbf{\dot{g}}$ denotes the time derivative of $\mathbf{g}$:
\begin{equation} \label{lemma}
\xi = \mathbf{J} \dot{\mathbf{q}} + \mathbf{C} \dot{\mathbf{w}}
\end{equation}

Now, consider $\mathbf{g}(s)$ and $\mathbf{b}$ as the solution to the IVP formed by (\ref{compact_odes})-(\ref{initial_cond}). Since they are fully determined by $\mathbf{x}$, their total derivatives consist only of their partial derivatives w.r.t. each component of $\mathbf{x}$. For $\mathbf{g}(s)$, since it stays on $SE(3)$, we consider the spatial twists given by \cite{MLS_robotics_txtbk}:

\begin{equation} \label{IVP_dg}
    \mathbf{E} = [\mathbf{E_q}~ \mathbf{E_w}~ \mathbf{E_u}] = \left[ \left ( (\partial_{x_1} \mathbf{g}) \textbf{g}^{-1} \right )^{\vee}~ ...~
    \left ( (\partial_{x_N} \mathbf{g}) \textbf{g}^{-1} \right )^{\vee} \right ] 
\end{equation}

And we can obtain $\mathbf{\xi}$ for the $\mathbf{g}(s)$ as an IVP solution by using the chain rule:
\begin{equation} \label{IVP_twist}
        \xi = \mathbf{E_q} \, \dot{\mathbf{q}} +
        \mathbf{E_w} \, \dot{\mathbf{w}} + 
        \mathbf{E_u} \, \dot{\mathbf{y}}_u(0)
\end{equation}

The partial derivatives of $\mathbf{b}$ are given by
\begin{equation} \label{IVP_db}
    \mathbf{B} = [\mathbf{B_q}~ \mathbf{B_w}~ \mathbf{B_u}] = \left[ \partial_{x_1} \mathbf{b} ~...~
    \partial_{x_N} \mathbf{b} \right ] 
\end{equation}

Observe that, for the real system, $\mathbf{g}(s)$ should always remain as a solution to the BVP (\ref{compact_odes})-(\ref{boundary_definition}) while varying with $\mathbf{q}$ and $\mathbf{w}$. This requires $\mathbf{y}_u(0)$ to vary in a way that it compensates the variations in $\mathbf{q}$ and $\mathbf{w}$, such that $\mathbf{b}(\mathbf{x})=\mathbf{0}$ always holds. This constraint is obtained by taking the time derivative of (\ref{boundary_definition}):
\begin{equation} \label{IVP_boundary_constr}
    \mathbf{0} = \Dot{\mathbf{b}} = \mathbf{B_q} \, \dot{\mathbf{q}} + 
        \mathbf{B_w} \, \dot{\mathbf{w}} + 
        \mathbf{B_u} \, \dot{\mathbf{y}}_u(0)
\end{equation}

Using (\ref{IVP_boundary_constr}) to eliminate the $\Dot{\mathbf{y}}_u(0)$ in (\ref{IVP_twist}) results in the expression of (\ref{lemma}) by partial derivatives:
\begin{equation}
    \mathbf{\xi} = (\mathbf{E_q} - \mathbf{E_u} \mathbf{B_u^{\dagger}} \mathbf{B_q})\Dot{\mathbf{q}} + (\mathbf{E_w} - \mathbf{E_u} \mathbf{B_u^{\dagger}} \mathbf{B_w})\Dot{\mathbf{w}}
\end{equation}
from which we obtain the Jacobian and compliance matrices:
\begin{equation} \label{Jacob_derive}
        \mathbf{J} = \mathbf{E_q} - \mathbf{E_u} \mathbf{B_u^{\dagger}} \mathbf{B_q},~~
        \mathbf{C} = \mathbf{E_w} - \mathbf{E_u} \mathbf{B_u^{\dagger}} \mathbf{B_w}
\end{equation}
where $\mathbf{B_u^{\dagger}}$ is the pseudo-inverse of $\mathbf{B_u}$.

For gradient-based redundancy resolution, we calculate the derivatives of Jacobian and compliance matrix w.r.t. $\mathbf{q}$, i.e. the Hessians, using the same technique. Treating $\mathbf{J}$ and $\mathbf{C}$ as functions of solutions to the IVP and taking the time derivatives yields:
\begin{equation*}
    \begin{aligned}
        \dot{\mathbf{J}} & = \partial_{\mathbf{q}} \mathbf{J} \otimes \dot{\mathbf{q}} + \partial_{\mathbf{w}} \mathbf{J} \otimes \dot{\mathbf{w}} + \partial_{\mathbf{u}} \mathbf{J} \otimes \dot{\mathbf{y}}_u(0) \\
        \dot{\mathbf{C}} & = \partial_{\mathbf{q}} \mathbf{C} \otimes \dot{\mathbf{q}} + \partial_{\mathbf{w}} \mathbf{C} \otimes \dot{\mathbf{w}} + \partial_{\mathbf{u}} \mathbf{C} \otimes \dot{\mathbf{y}}_u(0) 
    \end{aligned}
\end{equation*}
Eliminating the $\dot{\mathbf{y}}_u(0)$ using (\ref{IVP_boundary_constr}), the derivatives of $\mathbf{J}$ and $\mathbf{C}$ that satisfy the BVP are given by:
\begin{equation} \label{hess_derive}
    \dot{\mathbf{J}} = \text{d}_{\mathbf{q}} \mathbf{J} \otimes \dot{\mathbf{q}} + \text{d}_{\mathbf{w}} \mathbf{J} \otimes \dot{\mathbf{w}},~~
    \dot{\mathbf{C}} = \text{d}_{\mathbf{q}} \mathbf{C} \otimes \dot{\mathbf{q}} + \text{d}_{\mathbf{w}} \mathbf{C} \otimes \dot{\mathbf{w}} 
\end{equation}
where the Hessians used for redundancy resolution are
\begin{equation} \label{grad_J_C}
    \text{d}_{\mathbf{q}} \mathbf{J} = \partial_{\mathbf{q}} \mathbf{J} - \partial_{\mathbf{u}} \mathbf{J} \otimes (\mathbf{B_u^{\dagger}} \mathbf{B_q}),~
    \text{d}_{\mathbf{q}} \mathbf{C} = \partial_{\mathbf{q}} \mathbf{C} - \partial_{\mathbf{u}} \mathbf{C} \otimes (\mathbf{B_u^{\dagger}} \mathbf{B_q})
    % &     \text{d}_{\mathbf{w}} \mathbf{J} = \partial_{\mathbf{w}} \mathbf{J} - \partial_{\mathbf{u}} \mathbf{J} \otimes (\mathbf{B_u^{\dagger}} \mathbf{B_w}),~
    % \text{d}_{\mathbf{w}} \mathbf{C} = \partial_{\mathbf{w}} \mathbf{C} - \partial_{\mathbf{u}} \mathbf{C} \otimes (\mathbf{B_u^{\dagger}} \mathbf{B_w})
\end{equation}
To obtain the partial derivatives in the above equations, we further define the derivatives of $\mathbf{E}$ and $\mathbf{B}$ %\yf{Be AWARE of what you are writing, it's B not V here. Please don't make me check all your math details again.} 
 as $ \mathbf{D} = \partial_\mathbf{x} \mathbf{E}$ and $ \mathbf{A} = \partial_\mathbf{x} \mathbf{B}$, and take the derivatives of (\ref{Jacob_derive}) w.r.t. $\mathbf{x}$:

\begin{equation} \label{hessian_intermedia}
    \begin{aligned}
    \partial_{x_{\text{r}}} \mathbf{J} 
    & = \mathbf{D}_{\textbf{q},\text{r}} - \mathbf{D}_{\textbf{u},\text{r}} \mathbf{B_u^{\dagger}} \mathbf{B_q} 
    - \mathbf{E_u} \mathbf{B_u^{\dagger}} \mathbf{A}_{\textbf{u},\text{r}} \mathbf{B_u^{\dagger}} \mathbf{B_q} - \mathbf{E_u} \mathbf{B_u^{\dagger}} \mathbf{A}_{\textbf{q},\text{r}} \\
    \partial_{x_{\text{r}}} \mathbf{C} 
    & = \mathbf{D}_{\textbf{w},\text{r}} - \mathbf{D}_{\textbf{u},\text{r}} \mathbf{B_u^{\dagger}} \mathbf{B_w} 
    - \mathbf{E_u} \mathbf{B_u^{\dagger}} \mathbf{A}_{\textbf{u},\text{r}} \mathbf{B_u^{\dagger}} \mathbf{B_w} - \mathbf{E_u} \mathbf{B_u^{\dagger}} \mathbf{A}_{\textbf{w},\text{r}}
    \end{aligned}
\end{equation}
where the $r$-th page of tensors $\mathbf{D}$ and $\mathbf{A}$ are denoted as 
\begin{equation}
    \begin{aligned}
        \mathbf{D}_{\text{r}} & = [ \mathbf{D}_{\textbf{q},\text{r}}~  \mathbf{D}_{\textbf{w},\text{r}}~  \mathbf{D}_{\textbf{u},\text{r}}] = [\partial_{x_{\text{r}}} \mathbf{E}_{\textbf{q}}~ \partial_{x_{\text{r}}} \mathbf{E}_{\textbf{w}}~ \partial_{x_{\text{r}}} \mathbf{E}_{\textbf{u}}] \\
        \mathbf{A}_{\text{r}} & = [ \mathbf{A}_{\textbf{q},\text{r}}~  \mathbf{A}_{\textbf{w},\text{r}}~  \mathbf{A}_{\textbf{u},\text{r}}] = [\partial_{x_{\text{r}}} \mathbf{B}_{\textbf{q}}~ \partial_{x_{\text{r}}} \mathbf{B}_{\textbf{w}}~ \partial_{x_{\text{r}}} \mathbf{B}_{\textbf{u}}] 
    \end{aligned}
\end{equation}

We can observe from (\ref{Jacob_derive}), (\ref{grad_J_C}) and (\ref{hessian_intermedia}) that, calculating the Jacobian and compliance matrices and the corresponding Hessians requires calculating $\mathbf{B}$, $\mathbf{E}$, $\mathbf{A}$, and $\mathbf{D}$. While $\mathbf{E}$ and $\mathbf{D}$ are derivatives of $\mathbf{g}(s)$ and can be obtained from initial conditions since they exist along the robot length, $\mathbf{B}$ and $\mathbf{A}$ are evaluated only at $s_{e,i}$ and cannot propagate with $s$. However, note from (\ref{boundary_definition}) that $\mathbf{b}$ is a function of $\mathbf{y}(s_{e,i})$ and $\mathbf{g}(s_{e,i})$, hence we can obtain $\mathbf{B}$ and $\mathbf{A}$ by propagating the derivatives of $\mathbf{y}(s)$ and $\mathbf{g}(s)$ w.r.t. $\mathbf{x}$. Denote the first- and second-order derivatives of the state vector $\mathbf{y}$ w.r.t. $\mathbf{x}$ as:
\begin{equation*} %\label{V_U_definition}
    \mathbf{V} = \partial_{\mathbf{x}} \mathbf{y}, \quad 
    \mathbf{U} =  \partial^2_{\mathbf{x}, \mathbf{x}} \mathbf{y}
\end{equation*}
Then $\textbf{B}$ and $\mathbf{A}$ can be obtained by taking derivatives of $\mathbf{b}(\mathbf{x})$ and plugging in values of $\mathbf{V}$ and $\mathbf{U}$ at $s_{e,i}$.
Note that the first-order partial derivatives $\mathbf{E}$, $\mathbf{V}$ and the second-order partial derivatives $\mathbf{D}$, $\mathbf{U}$
are themselves functions of $s$, they can be calculated by integrating along the arc length through a new set of ODEs. Since $\mathbf{g}$ and $\mathbf{y}$ are piecewise continously differentiable, we have the relationship $\mathbf{V}^{\prime} = (\partial_{\mathbf{x}} \mathbf{y})^{\prime} = \partial_{\mathbf{x}} (\mathbf{y}^{\prime})$. Hence the $k$-th column of matrices $\mathbf{E}^\prime$ and $\mathbf{V}^\prime$ are given by:
\begin{equation}\label{EV}
\begin{aligned}
    \mathbf{E}_k^{\prime} & = \left( \partial_{x_k} (\mathbf{g}^{\prime}) \cdot \mathbf{g}^{-1} + \partial_{x_k} \mathbf{g} \cdot (\mathbf{g}^{-1})^{\prime} \right)^{\vee} = \left( \mathbf{g} \cdot \partial_{x_k} \mathbf{\hat{\zeta}} \cdot \mathbf{g}^{-1} \right)^{\vee} \\
    \mathbf{V}_k^{\prime} & =  \partial_{\mathbf{y}} \mathbf{f} \cdot \mathbf{V}_k + 
    \partial_{x_k} \mathbf{f} + \partial_{\text{vec}(\textbf{R})} \mathbf{f} \cdot \text{vec} (\partial_{x_k} \mathbf{R})
\end{aligned}
\end{equation}
where the partial derivative of $\zeta$ and $\mathbf{R}$ can be obtained by $\partial_{x_k} \mathbf{\zeta} = \partial_{\mathbf{y}} \zeta \cdot \textbf{V}_k $, $ \partial_{x_k} \mathbf{R} = ([\mathbf{0}_{3\times 3} ~ \mathbf{I}_{3 \times 3}] \mathbf{E}_k )^{\vee} \mathbf{R}$, and $\text{vec}()$ reshapes a matrix into a column vector. Denoting (\ref{EV}) as $[(\mathbf{E}^\prime)^T~(\mathbf{V}^\prime)^T]^T = \mathbf{f}^1(s,\mathbf{y},\mathbf{R},\mathbf{w},\mathbf{E},\mathbf{V})$, the ODEs for the second-order derivatives are derived by taking the derivative of $\mathbf{f}^1$ w.r.t. $\mathbf{x}$. The $r$-th page of tensors $\mathbf{D}^\prime$, $\mathbf{U}^\prime$ is given by:
\begin{equation}\label{DU}
    \begin{aligned}
        \begin{bmatrix}
            \mathbf{D}_{\text{r}} \\ \mathbf{U}_{\text{r}}
        \end{bmatrix}^{\prime} & = \partial_{\mathbf{y}} \mathbf{f}^1 \cdot \mathbf{V}_r + \partial_{x_r} \mathbf{f}^1 +  \partial_{\text{vec}(\textbf{R})} \mathbf{f}^1 \cdot \text{vec} (\partial_{x_k} \mathbf{R})  \\ & + \partial_{\text{vec}(\textbf{V})} \mathbf{f}^1 \cdot \text{vec} (\mathbf{U}_{\text{r}}) + \partial_{\text{vec}(\textbf{E})} \mathbf{f}^1 \cdot \text{vec} (\mathbf{D}_{\text{r}})
    \end{aligned}
\end{equation}

\begin{figure}[t!]
    \centering
    \includegraphics[width = 1.00\linewidth]{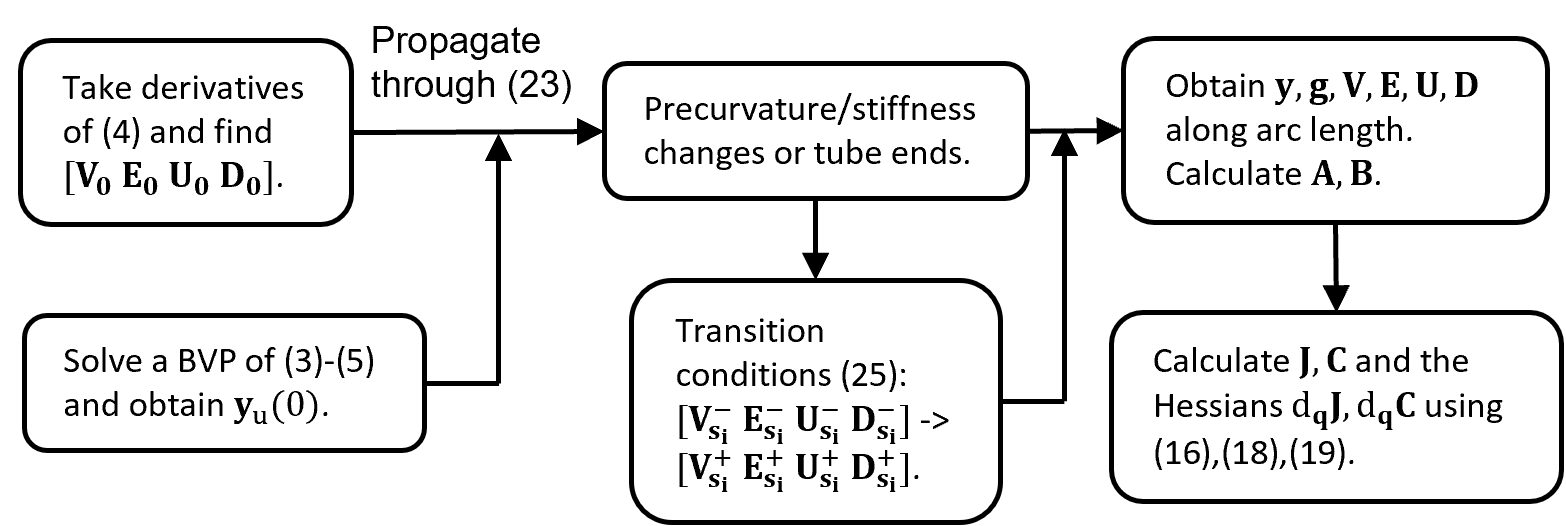}
    \caption{Flow chart of steps for calculating the Hessians via derivative propagation.} 
    \label{fig:workflow}
    \vspace{- 2mm}
\end{figure}

 Combining (\ref{EV}) and (\ref{DU}) with (1) gives an augmented system of ODEs: 
 % \yf{Probably need a flow chart to summarize the entire process. It's pretty hard to follow with only text descriptions.}
% \yf{explain in earlier paragraphs that a BVP is first solved for forward kinematics, and then this IVP is solved for Hessian}
\begin{subequations} \label{ode_hess}
    \begin{align}
        \mathbf{g}^\prime(s) \,\,\,\, &= \mathbf{g}\Hat{\mathbf{\zeta}}(\mathbf{y}) \label{ode_hess_1} \\
        \mathbf{y}^{\prime}(s) \,\,\,\, & = \mathbf{f}(s, \mathbf{y},\mathbf{R}, \mathbf{w}) \label{ode_hess_2} \\
        % \mathbf{E}^{\prime} & = \begin{bmatrix}
        %     (\partial_{\mathbf{x}} \mathbf{R}) \mathbf{e}_3 \\ \mathbf{R} (\partial_{\mathbf{x}} \mathbf{u}_1) \end{bmatrix} \\
        % \mathbf{D}^{\prime} & = \begin{bmatrix}
        %     (\partial^2_{\mathbf{x},\mathbf{x}} \mathbf{R}) \mathbf{e}_3 \\ (\partial_{\mathbf{x}} \mathbf{R}) (\partial_{\mathbf{x}} \mathbf{u}_1) + \mathbf{R} (\partial^2_{\mathbf{x},\mathbf{x}} \mathbf{u}_1) \end{bmatrix} \\
        {\begin{bmatrix}
            \mathbf{E}(s) \\ \mathbf{V}(s)
        \end{bmatrix}}^{\prime} & = \mathbf{f}^1(s,\mathbf{y}, \mathbf{R}, \mathbf{w}, \mathbf{E},\mathbf{V}) \label{ode_hess_3} \\
        {\begin{bmatrix}
            \mathbf{D}(s) \\ \mathbf{U}(s)
        \end{bmatrix}}^{\prime} & = \textbf{f}^2(s,\mathbf{y}, \mathbf{R}, \mathbf{w}, \mathbf{E},\mathbf{V}, \mathbf{D},\mathbf{U}) \label{ode_hess_4}
    \end{align}
\end{subequations}
where (\ref{ode_hess_4}) is the concise form of (\ref{DU}).
Using initial values calculated by taking the first- and second-order derivatives of (\ref{initial_cond}) w.r.t. $\mathbf{x}$, (\ref{ode_hess}) can be solved as an IVP.

Note that at the end of each tube or at positions where the precurvature or stiffness of tubes has discontinuity, partial derivatives $\mathbf{V}$, $\mathbf{E}$, $\mathbf{U}$, $\mathbf{D}$ are discontinuous since they are continuous functions of stiffness and precurvature. Their transition functions at these positions can be obtained by taking the derivatives of the transition functions of $\mathbf{y}$ and $\mathbf{g}$. The arc length when these discontinuities appear is denoted as $s_i$, representing either $s_{e,i}$ or $s_{t,i}$. Transitions of $\mathbf{y}$ and $\mathbf{g}$ are given by:
\begin{subequations} \label{transition}
\begin{align}
    \mathbf{y}^{+}(s, \mathbf{x})|_{s=s_i} & = \mathbf{h}(\mathbf{y}^{-}(s, \mathbf{x}))|_{s=s_i}\label{transition_y} \\
    \mathbf{g}^{+}(s, \mathbf{x})|_{s=s_i} & = \mathbf{g}^{-}(s, \mathbf{x})|_{s=s_i} \label{transition_g}
\end{align}
\end{subequations}
where $-$ and $+$ denote the state vector immediately before and after the transition point, respectively, and $\mathbf{h}()$ is the transition function of $\mathbf{y}$ that enforces boundary conditions (\ref{boundary_definition}) and static equilibrium at transition points $s_i$. Define the elements in vector $\mathbf{x}$ excluding $\beta_i$ as $\phi_j \in \mathbf{x} \backslash {\beta_i}$. To take the first- and second-order derivatives of equation (\ref{transition_y}) over $\beta_i$ and $\phi_j$, we note that $s_i$ is a linear function of $\beta_i$, hence $\mathrm{d}_{\beta_i} \textbf{y}|_{s=s_{e,i}} = \partial_{\beta_i} \textbf{y} + \partial_{s} \textbf{y}\cdot\mathrm{d}_{\beta_i}s_{e,i} = \partial_{\beta_i} \textbf{y} + \partial_{s} \textbf{y}$ and we have
\begin{equation}
    \begin{aligned}
    % equation 0
    \partial_{\phi_j} {\mathbf{y}}^{+} & = \partial_{\mathbf{y}} \mathbf{h} ( \partial_{\phi_j} {\mathbf{y}}^{-}), \\
    % equation 1
        {\partial_{\beta_i} \textbf{y}}^{+} & = 
        \partial_{\mathbf{y}} \textbf{h}({\partial_{\beta_i} \textbf{y}}^{-} + 
       \partial_{s} \textbf{y}^{-}) - {\partial_{s} \textbf{y}}^{+}, \\
    % equation 2
        {\partial^2_{\beta_i,\phi_j} \textbf{y}}^{+} & = 
        \partial^2_{\mathbf{y},\mathbf{y}} \textbf{h}({\partial^2_{\beta_i,\phi_j} \mathbf{y}}^{-} + 
       {\partial^2_{s,\phi_j} \mathbf{y}}^{-}) - {\partial^2_{s,\phi_j} \textbf{y}}^{+}, \\
    % equation 3
        \partial^2_{\beta_i,\beta_i} \textbf{y}^{+}  & =
        \partial^2_{\mathbf{y},\mathbf{y}} \textbf{h}(\partial^2_{\beta_i,\beta_i} \textbf{y}^{-} + 
        2 \partial^2_{s,\beta_i} \textbf{y}^{-} + \partial^2_{s,s} \textbf{y}^{-}) \\
        & \quad - 2 \partial^2_{s,\beta_i} \textbf{y}^{+} - \partial^2_{s,s} \textbf{y}^{+}
    \end{aligned}
\end{equation}

The transition conditions for derivatives of $\mathbf{g}$ takes a similar form. When the forward integration of (\ref{ode_hess}) passes through transition points, the above conditions are used to properly transition the augmented state variables in (\ref{ode_hess}).

By solving the IVP (\ref{ode_hess}), we obtain $\textbf{V},\textbf{E},\textbf{U},\textbf{D}$, from which we can calculate $\mathbf{B}$, $\mathbf{A}$. Then $\textbf{B},\textbf{A},\textbf{E},\textbf{D}$ are plugged into (\ref{Jacob_derive}), (\ref{hessian_intermedia}), and (\ref{grad_J_C}) to obtain $\textbf{J}$, $\textbf{C}$ and their gradients. While the finite difference method requires solving several BVPs or IVPs, the derivative propagation method only needs to solve an augmented IVP and therefore reduces the computational load. The overall procedure of calculating the Hessians is summarized in Fig.\ref{fig:workflow}.

\section{Task-Specific Redundancy Resolution} \label{section_controller}

In this section, we present the redundancy resolution scheme to regulate the robot configuration for effective trajectory tracking. We incorporate performance index optimization in the redundancy resolution for tasks including singularity avoidance, obstacle avoidance, and tracking under external force. The redundancy resolution proposed in this section can be used as low-level building blocks for a high-level task and motion planner.  

\subsection{Trajectory Tracking and Joint Limit Avoidance}\label{track_joint_limit}
 The primary task in all scenarios we consider is tracking a desired trajectory designated by either teleoperation or a motion planner, while avoiding joint limits. Consider points of interest on the robot where desired twists $\mathbf{\xi}_{j,d},~j=1,...,m$ are designated. 
The primary task can be formulated as:
\begin{subequations} \label{primary_opt}
\begin{align}
   & \quad \min_{\Dot{\mathbf{q}}} \quad \dot{\mathbf{q}}^T \mathbf{W}(\mathbf{q}) \dot{\mathbf{q}}\label{primary_cost} \\
    & \quad \text{s.t.} \quad \quad \mathbf{J}_j \, \dot{\mathbf{q}} = \mathbf{\xi}_{j,d}, \quad j = 1,...,m\label{primary_constr}
\end{align}
\end{subequations}
where $\mathbf{J}_{j}$ is the Jacobian of the $j$th point of interest, $\mathbf{W}$ is an adaptive positive definite weight matrix.

The cost function (\ref{primary_cost}) is designed to penalize the velocity that drives joints to their limits. The joint limit of CTR consists of limits on the exposed lengths $\gamma_i$ (Fig.\ref{fig:CTR_illustration}), which prevent withdrawing the inner tube entirely into the outer tube ($\gamma_i>0$) or extending the inner tube too much ($\beta_i<\beta_{i+1}$).
Hence the weight matrix is defined as $ \mathbf{W}(\mathbf{q}) = \text{diag} \left ([1,w_1,1,w_2,1,w_3] \right )$ so that only $\beta_i$ is  regulated. To penalize the joint velocity that drives the exposed length $\gamma_i$ to its limit, the adaptive weight is defined as \cite{jointlimit_cost}:
\begin{equation}
    w_i = 1 + \left | \frac{(\gamma_{i,max} - \gamma_{i,min})^2 (\gamma_i - \bar{\gamma}_i)}{2 (\gamma_{i,max} - \gamma_i)^2 (\gamma_i - \gamma_{i,min})^2} \right |
\end{equation}
where $\gamma_i \in \left [\gamma_{i,min}, \gamma_{i,max} \right ]$, and $\bar{\gamma}_i = (\gamma_{i,min}+ \gamma_{i,max}) / 2$. When $\gamma_i$ approaches the limits, the weight $w_i$ approaches infinity and penalizes $\dot{\gamma}_i$ towards 0.

To solve (\ref{primary_opt}), define the augmented Jacobian \cite{Augmented_Jacobian_Egeland} as $\mathbf{J} = [\mathbf{J}_1^T, ...,\mathbf{J}_m^T]^T$, and the augmented desired twist $\mathbf{\xi}_d = [\mathbf{\xi}_{1,d}^T, ..., \mathbf{\xi}_{m,d}^T]^T $. The transformation $\mathbf{J}_w = \mathbf{J} \mathbf{W} ^{-\frac{1}{2}}$, $\dot{\mathbf{q}}_w = \mathbf{W} ^ {\frac{1}{2}} \dot{\mathbf{q}}$ simplifies (\ref{primary_opt}) into minimizing $||\dot{\mathbf{q}}_w||^2$ while satisfying $\mathbf{J}_w\Dot{\mathbf{q}}_w=\mathbf{\xi}_d$, which has the closed-form solution 
\begin{equation} \label{primary_task}
    \dot{\mathbf{q}} = \mathbf{W}^{-\frac{1}{2}}\mathbf{J}_w^{\dagger} \mathbf{\xi}_d = \mathbf{W}^{-1}\mathbf{J}^T(\mathbf{J}\mathbf{W}^{-1}\mathbf{J}^T)^{-1}\mathbf{\xi}_d
\end{equation}
where $\mathbf{J}^{\dagger} = \mathbf{J}^{T} \left (\mathbf{J} \mathbf{J}^{T} \right ) ^{-1}$ for a redundant robot.

\subsection{Task-Specific Performance Index} \label{performance_index}
We consider three scenarios and derive the task-specific performance indices together with their gradient based on previous derivations.

    \textbf{\textit{Scenario 1}}: When the robot is tracking a desired trajectory in free space, it is beneficial to maintain a relatively high VMI to avoid singular configuration. 
    Particularly, if the desired trajectory is suddenly updated, a sufficient VMI will enable the robot to follow the new trajectory immediately. It is also reported that increasing VMI helps avoid the snapping of the CTR \cite{khadem2019RR_manipul}.
    Here, we consider the VMI of robot tip and incorporate the joint limit into the VMI by substituting the original Jacobian in (\ref{velo_manipul_def}) with the weighted Jacobian $\mathbf{J}_w$ defined earlier. Note that when $\beta_i$ approaches its limit, the corresponding column in $\mathbf{J}_w$ is penalized towards $\mathbf{0}$, effectively reducing the manipulability generated by $\beta_i$. 
    For gradient-based redundancy resolution, the analytical gradient of the VMI can be derived using Jacobi's formula:
\begin{equation}
    \partial_{q_i} \mathbf{\mu}_v(\mathbf{q}) = \frac{1}{2} \text{det}(\bm{\Gamma}_v)^{-\frac{1}{2}} \text{det}(\bm{\Gamma}_v) \text{trace}(\bm{\Gamma}_v^{-1} \partial_{q_i} \bm{\Gamma}_v)
\end{equation}
where $\bm{\Gamma}_v = \mathbf{J}_w \mathbf{J}_w^T$, and the partial derivative $\partial_{q_i} \bm{\Gamma}_v$ :
\begin{equation}
    \partial_{q_i} \bm{\Gamma}_v = (\partial_{q_i} \mathbf{J}) \mathbf{W}^{-1} \mathbf{J}^T  + \mathbf{J} \mathbf{W}^{-1} (\partial_{q_i} \mathbf{J})^T
\end{equation}
can be obtained by using the Hessian from (\ref{grad_J_C}).  
\textbf{\textit{Scenario 2}}: When navigating through a confined space, the CTR needs to avoid obstacles that can potentially collide with robot body. Increasing the robot body VMI would improve the motion capability of the robot and facilitate obstacle avoidance. We assume that a map of obstacles is known, and the points of interest on CTR can be determined by task-specific criteria, such as selecting the closest point to each nearby obstacle. At each point of interest, a desired velocity that guides the robot away from obstacles can be obtained.  Denote the unit vector of the desired direction as $\mathbf{\rho}$, an oriented VMI is defined as  the projection of VME along $\mathbf{\rho}$, and is obtained by: 
\begin{equation}
    \label{oriented_VMI}    \mathbf{\mu}_{j,v}^d(\mathbf{q},\mathbf{\rho}) = \left [ \mathbf{\rho}^T (\mathbf{J}_j \mathbf{J}^T_j)^{-1} \mathbf{\rho} \right ]^{-\frac{1}{2}}
\end{equation}
A weighted whole-body VMI is then defined as:
\begin{equation} \label{body_manipul}
    \mathbf{\mu}_o(\mathbf{q}) = \sum_{j=1}^p c_j \mathbf{\mu}_{j,v}^d(\mathbf{q},\mathbf{\rho}_j)
\end{equation}
where $\mathbf{\mu}_{j,v}^d$ is the oriented VMI of the $j$-th point of interest. The weight $c_j$ describes the relative importance of the $j$-th point of interest and can be determined as a function of e.g. the distance between the robot body.
The gradient of  $\mathbf{\mu}_c^d$ is derived as: 
\begin{equation}\label{grad_mu}
     \partial_{q_i} \mathbf{\mu}^d_{j,v}(\mathbf{q},\mathbf{\rho}) = \frac{1}{2} (\mathbf{\rho}^T \bm{\Gamma}_{j,v}^{-1} \mathbf{\rho})^{-\frac{3}{2}} \mathbf{\rho}^T \bm{\Gamma}_{j,v}^{-1} (\partial_{q_i} \bm{\Gamma}_{j,v}) \bm{\Gamma}_{j,v}^{-1} \mathbf{\rho}
\end{equation}
where $\bm{\Gamma}_{j,v} = \mathbf{J}_j \mathbf{J}^T_j$, and the partial derivative $\partial_{q_i} \bm{\Gamma}_v$ can be obtained using the Hessian in equation (\ref{grad_J_C})

\textbf{\textit{Scenario 3}}: When performing certain surgical procedures, such as suturing or forcep-based biopsy, there is a concentrated external load $\mathbf{w}$ applied to the robot tip that deforms the robot, potentially leading to undesired behavior. 
It is usually desirable to suppress the robot  deformation while following the designated trajectory. This can be achieved by minimizing the compliance in the direction of the external load. Similar to the definition of the oriented VMI, we denote the unit vector along the direction of the tip load as $\mathbf{\nu}$, and the compliance in this direction is obtained by the projection of CME along $\mathbf{\nu}$:
\begin{equation}
    \label{compliance_definition}
    \mathbf{\mu}_c^d(\mathbf{q},\mathbf{\nu}) = \left [ \mathbf{\nu}^T (\mathbf{C} \mathbf{C}^T)^{-1} \mathbf{\nu} \right ]^{-\frac{1}{2}}
\end{equation}
Its gradient $\partial_{q_i} \mathbf{\mu}^d_c(\mathbf{q},\mathbf{\nu})$ can be obtained similar to (\ref{grad_mu}).

\subsection{Redundancy Resolution with Task-Specific Gradient Projection (TSGP)}
The redundancy resolution is conducted such that the trajectory tracking and joint limit avoidance (\ref{primary_task}) is firstly satisfied, then the gradient-based optimization of the task-specific performance index is performed using the remaining degrees of freedom. 
This is achieved by projecting the gradient of the performance index $\mu$ into the null-space of $\textbf{J}_w$:
\begin{equation} 
    \dot{\mathbf{q}}_w = \mathbf{J}_w^{\dagger} \mathbf{\xi}_d + \alpha (\mathbf{I} - \mathbf{J}_w^{\dagger} \mathbf{J}_w) \nabla \mu
\end{equation}
where $\alpha$ is a scalar parameter. A positive $\alpha$ would increase $\mu$, and a negative $\alpha$ would decrease it. As mentioned in \cite{jointlimit_cost}, choosing a suitable gain $\alpha$ across the whole workspace is critical for TSGP. 

In this paper, we adopt the method in our recent work \cite{azizkhani2023design} to find a suitable $\alpha$ that balances the desired velocity and null-space projection for manipulability optimization. 

The final instantaneous joint velocity of the TSGP controller is given by
\begin{equation} \label{GPM_equation}
\begin{aligned}
    \dot{\mathbf{q}} & = \mathbf{W}^{-1} \mathbf{J}^T (\mathbf{J} \mathbf{W}^{-1} \mathbf{J}^T )^{-1} \mathbf{\xi}_d  \\ &  \quad + \alpha \left ( \mathbf{I} - \mathbf{W}^{-1} \mathbf{J}^T (\mathbf{J}\mathbf{W}^{-1} \mathbf{J}^T )^{-1} \mathbf{J} \right ) \mathbf{W}^{-1} \nabla \mu
\end{aligned}
\end{equation}

\section{Simulation Study} \label{simulation}

To evaluate the performance of the algorithms developed in section \ref{Hessian} and \ref{section_controller}, trajectory tracking simulations were conducted using a 3-tube CTR to achieve tasks that reflect the scenarios presented in the previous section. 
Parameters of the CTR are given in Table \ref{tube_param}.  
We compared the performance of the proposed TSGP controller with two other widely used kinematic controllers, namely: 

\begin{enumerate}

\item The standard resolved-rates (RR) controller given by (\ref{primary_task}).

\item A generalized damped least-square (DLS) controller.
\end{enumerate}

The DLS controller tries to minimize a quadratic cost function \cite{burgner2013telerobotic}, \cite{AndersonRedundancyICRA17}:
\begin{equation}
    h(\Dot{\mathbf{q}}) = (\mathbf{J} \dot{\mathbf{q}} - \dot{\mathbf{p}}_d)^T \mathbf{W}_t (\mathbf{J} \dot{\mathbf{q}} - \dot{\mathbf{p}}_d) + \dot{\mathbf{q}}^T (\mathbf{W}_D + \mathbf{W}_J) \dot{\mathbf{q}}
\end{equation}
where $\mathbf{W}_t$, $\mathbf{W}_D$, $\mathbf{W}_J$ denotes the weight matrix for trajectory tracking, singularity robustness, and joint limit avoidance, respectively. The instantaneous joint velocity can be obtained by setting $\nabla h = \mathbf{0}$:
\begin{equation} \label{DLS_controller}
    \dot{\mathbf{q}}  = \left ( \mathbf{J}^T \mathbf{W}_t \mathbf{J} + \mathbf{W}_D + \mathbf{W}_J \right )^{-1} \mathbf{J}^T \mathbf{W}_t \Dot{\mathbf{p}}_d
\end{equation}
In this simulation, we used the $3\times6$ Jacobian for the desired linear velocity $\Dot{\mathbf{p}}_d$, and the DLS parameters were set to:
$
    \mathbf{W}_t = \mathbf{I}_{3\times3},~ \mathbf{W}_D = 0.001 \, \mathbf{I}_{6\times6},~ \mathbf{W}_J = 0.001 \, \mathbf{W}(\mathbf{q})
$. 
Here, $\mathbf{W}(\mathbf{q})$ is the joint limit weight defined in section \ref{track_joint_limit}. All algorithms are implemented in Matlab and run on a 8-core 2.30 GHz processor.

\begin{table}[t!]
\renewcommand{\arraystretch}{1.3}
\caption{Simulation Parameters for Tubes}
\label{tube_param}
\centering
\begin{tabular}{|l|l|l|l|}
\hline
 & Tube 1  & Tube 2  & Tube 3 \\
\hline
Inner Diameter (mm) & 0.640 & 0.953 & 1.400\\
Outer Diameter (mm) & 0.840 & 1.270 & 1.600\\
Straight Section Length (mm) & 500 & 250 & 100\\
Curved Section Length (mm) & 40 & 50 & 50\\
Curvature (m$^{-1}$) & 20 & 10 & 5\\
Young's Modulus, $E$ (GPa) & 60 & 60 & 60\\
Shear Modulus, $G$ (GPa) & 23.1 & 23.1 & 23.1\\
Joint Limit, $\gamma_{i,min}$ (mm)  & 10  &10  & 10 \\
Joint Limit, $\gamma_{i,max}$ (mm) & 200 & 200 & 200 \\
\hline
\end{tabular}
\vspace{- 2mm}
\end{table}
 
\subsection{Computational Efficiency for Hessian Calculation}
The computational efficiency of the derivative propagation method is evaluated by randomly sampling 10,000 
 configurations and calculating the corresponding Hessian. Different non-stiff ODE solvers provided by Matlab are explored for accuracy and efficiency. We compare the performance of the proposed method to that of the finite difference method, which calculates $\mathbf{D}$ and $\mathbf{U}$ by applying perturbations to $\mathbf{q}$ and calculating multiple $\mathbf{E}$s and $\mathbf{V}$s, each from one IVP using (\ref{ode_hess_1})-(\ref{ode_hess_3}). The Hessian calculated by the Runge-Kuta (4,5) solver (ode45) using finite difference is used as the reference to evaluate the errors of the derivative propagation. We record the largest relative error among all elements compared to the reference Hessians. Since the step length of different solvers vary, we evaluate the average CPU time together with the number of calls for the forward integration of (\ref{ode_hess}) for the derivative propagation method or equations (\ref{ode_hess_1})-(\ref{ode_hess_3}) for the finite difference method. We do not test calculating $\mathbf{E}$ and $\mathbf{V}$ using finite difference since this is significantly slower.

As shown in table \ref{Hessian_calc}, the Hessians by the proposed method agree with those of the finite difference method very well. The proposed method reduces the CPU times by 68\% and the number of calls of ODEs by 94\% compared to the finite difference method. In this comparison, we do not consider parallel computing. While it is straightforward to accelerate the finite difference by parallel computing, we note that the calculation of IVP can also be parallelized which can be used to further accelerate the derivative propagation method. In our simulation, finite difference using parallel computing typically results in $0.1$s CPU time, while 3 times speedup of ODE computing for $8$ threads is reported in literature \cite{ketcheson2014comparison_ode}. Therefore, we argue that the proposed method is more efficient given the same amount of computational resources.

\begin{table}[t!]
\renewcommand{\arraystretch}{1.3}
\caption{Computational Efficiency for the Hessian}
\label{Hessian_calc}
\centering
\begin{tabular}{|l|c|c|c|c|c|c|}
\hline
 & \multicolumn{3}{c|}{Finite Difference} & \multicolumn{3}{c|}{Derivative Propagation}\\
 \hline
ODE solver & ode45 & ode23 & ode113 & ode45 & ode23  & ode113 \\
\hline
Time (s)   & 0.498 & 0.320 & 0.293 & 0.159 & 0.133 & 0.125  \\
 PDE calls & 1645   &  908  & 953   &  95   &  67  &  54 \\
 Error (\%) & 0   &  3.82  & 1.90   &  0.0526   &  1.23  &  0.649 \\
\hline
\end{tabular}
\vspace{- 2mm}
\end{table}

\begin{figure*}[t!]
    \centering
    \includegraphics[width = 1.00\linewidth]{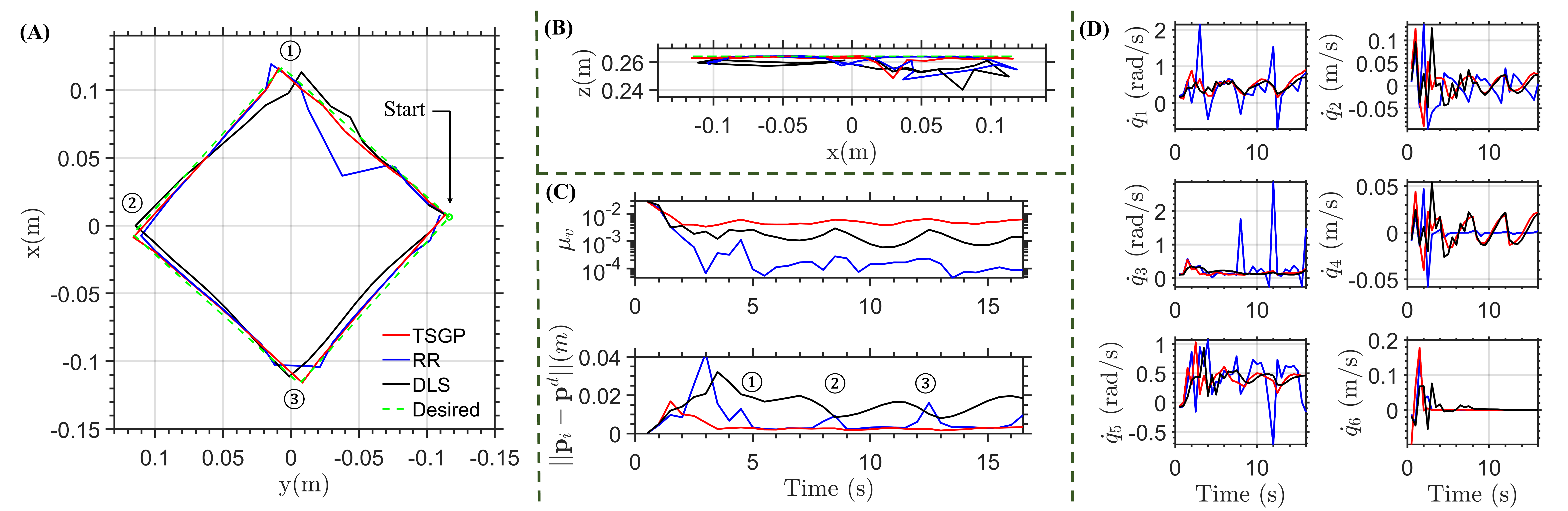}
    \vspace{- 6mm}
    \caption{Simulation results of tracking a square trajectory in free space.  
    (A) Top view of the trajectories. 
    (B) Side view of the trajectories. (C) The changes of VMIs and position errors along the trajectory.
    (D) The changes of joint velocities along the trajectory.} 
    \label{fig:GPM_velo_traj_follow}
    \vspace{- 2mm}
\end{figure*}

 \subsection{Free Space Trajectory Tracking} \label{velo_manipul}

For the scenario 1 in section  \ref{performance_index}, we used a square trajectory that contains $90^{\circ}$ turns and passes through neighborhoods of singularities. For each point along the trajectory, the controllers are given one iteration (0.5s time interval) to move the robot toward that point. This setup requires the robot to maintain a relatively high VMI to avoid singularities and follow the trajectory closely at sharp turns. Thus, the TSGP tried to maximize the VMI for this scenario.

Fig.\ref{fig:GPM_velo_traj_follow} gives the simulation results. 
As shown in Fig.\ref{fig:GPM_velo_traj_follow}-(C),  the VMI generated by TSGP is much higher than  RR and DLS throughout the trajectory, leading to  reduced tracking error. 
The RR controller results in large position errors at 
corners of the trajectory, since the robot has low VMIs at these points and cannot generate large enough velocity in the desired direction. The low $\mu_v$ generated by RR as in Fig.\ref{fig:GPM_velo_traj_follow}-(C) shows that the robot is close to singularities, which is also reflected in Fig.\ref{fig:GPM_velo_traj_follow}-(D) by large joint velocities generated by RR due to low motion capabilities. On the other hand, the DLS controller achieved a relatively higher VMI than RR and hence fewer spikes in joint velocities. However, this comes with the cost of an overall higher position error, since the control law (\ref{DLS_controller}) effectively damps the singular values of the Jacobian and distorts it.
 The TSGP controller avoids this issue by performing gradient ascent for VMI in the null-space of the Jacobian, which preserves the accuracy of the trajectory tracking task while keeping future Jacobians away from singularities.

\subsection{Obstacle Avoidance} \label{obstacle_avoid}
Corresponding to scenario 2 in \ref{performance_index}, the robot needs to achieve online obstacle avoidance while tracking a straight trajectory in this simulation. Apart from the robot tip, the point of interest is defined as the closest point on the robot curve to the obstacle. This point is updated during each iteration by performing a $k$-nearest neighbor (KNN) search between the discretized robot curve and the point cloud representing the surface of the obstacle.
We define the unit vector that is aligned with the closest point-pair and points towards the robot as $\mathbf{k}$.
Once the shortest distance is below a threshold, the following obstacle avoidance task is added to (\ref{primary_constr}):
\begin{equation}
    v = \mathbf{\rho}^T\mathbf{J}_v\Dot{\mathbf{q}},~~0 = \mathbf{k}^T\mathbf{J}_v\Dot{\mathbf{q}}
\end{equation}
\begin{figure}[t!]
    \centering
    \includegraphics[width = 1.00\linewidth]{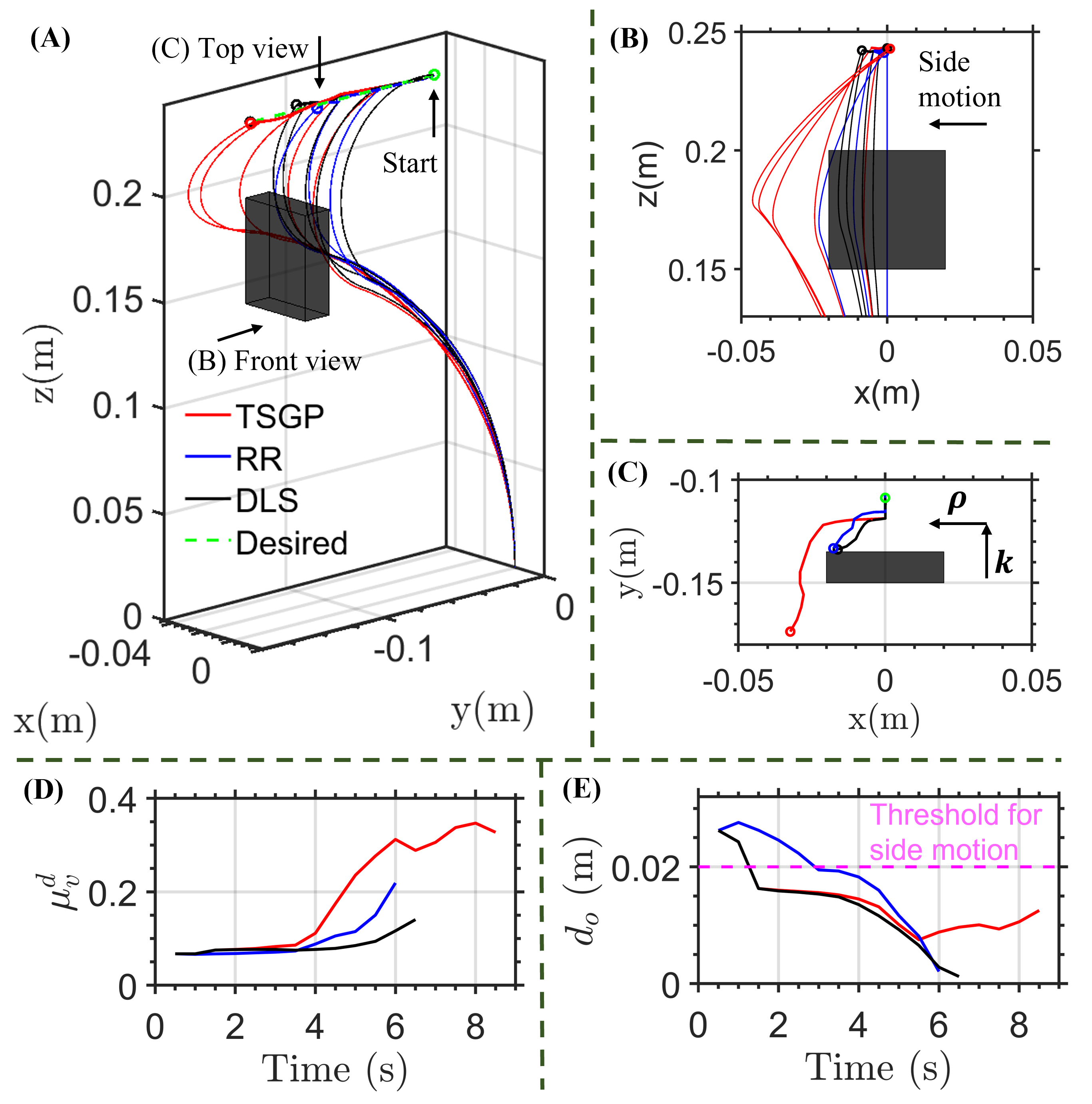}
    \vspace{- 6mm}
    \caption{Simulation results of obstacle avoidance. (A) Motion history of the robot using different controllers. (B) A front view of (A) showing the trajectories of the CTR shapes.  
    (C) A top view of (A) showing the trajectories of the closest point on the CTR to the obstacle. 
    (D) The changes of oriented VMIs along the trajectory. (E) The changes of the robot-obstacle distance along the trajectory.
    } 
    \label{fig:obstacle_avoid}
    \vspace{- 2mm}
\end{figure}
where $\mathbf{\rho} = (\mathbf{Re}_3)\times\mathbf{k}$, $\mathbf{Re}_3$ is the tangent vector of the robot at the point of interest, $v$ is the magnitude of the obstacle avoiding velocity, and $\mathbf{J}_v$ is the linear velocity Jacobian at the point closest to the obstacle.
For the 6-DoF CTR, the above task takes 2 DoFs and trajectory tracking takes 3 DoFs, hence 1 DoF is left to optimize the body manipulability in (\ref{body_manipul}). 

As shown in Fig.\ref{fig:obstacle_avoid}, a cuboid obstacle is placed to block the robot. Fig.\ref{fig:obstacle_avoid}-(C) shows the desired velocity direction that generates a sideward motion of the robot. The TSGP is capable of increasing the VMI of the robot at the point of interest, thus generating enough sideward motion to avoid the obstacle while following the tip trajectory. However, both the RR controller and DLS controller failed to avoid the obstacle due to the lack of sideward motion capability, as shown in Fig.\ref{fig:obstacle_avoid}-(D).

\subsection{Trajectory Tracking under External Load} \label{force_manipul}
The last simulation corresponds to the scenario 3 in \ref{performance_index}. While tracking a straight trajectory, a constant vertical force $\mathbf{F} = [0~0~-0.25 \text{N}]^T$ is applied to the tip of the CTR, and the TSGP controller minimizes the compliance in the vertical direction, i.e. $\mathbf{\mu}^d_c(\mathbf{q},-\mathbf{e}_3)$ defined in (\ref{compliance_definition}), to compensate for the effect of the external force.

\begin{figure}[t!]
    \centering
    \includegraphics[width = 1.00\linewidth]{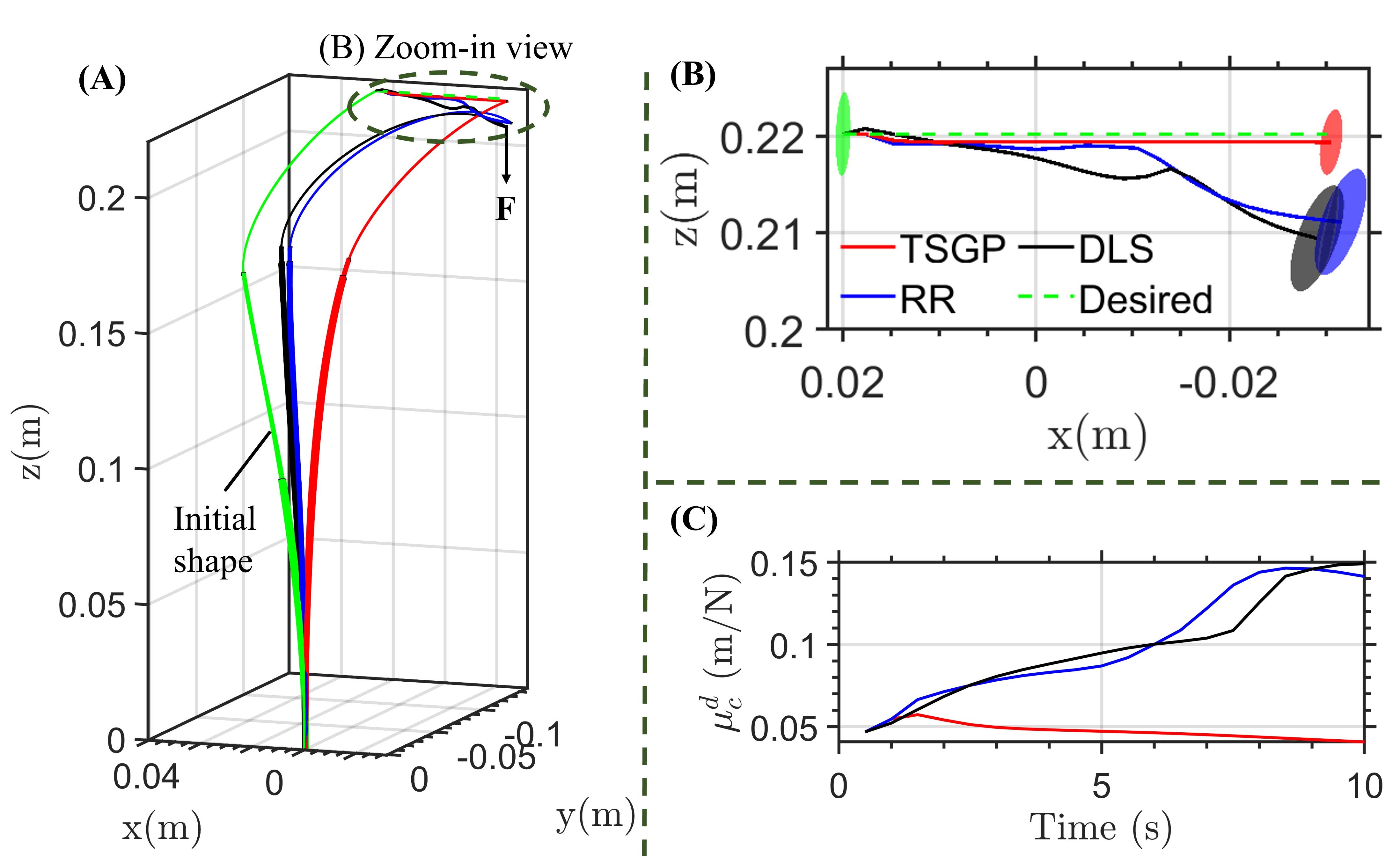}
    \vspace{- 6mm}
    \caption{Simulation results of trajectory tracking under vertical external force. % \yf{show the force in the figure} 
    (A) Motion histories of the robot using different controllers. (B) Zoom-in view of the robot tip trajectories and the CMEs of different controllers at the last time step. (C) The changes of oriented CMIs along the trajectory.
    }
    \label{fig:stiffness_control_force}
    \vspace{- 2mm}
\end{figure}

As shown in Fig.\ref{fig:stiffness_control_force}-(A) and (B),  
% \yc{please correct this}
the position error of the TSGP controller is significantly lower than those using RR and DLS. It can be observed from Fig.\ref{fig:stiffness_control_force}-(A) that the robot shapes for RR and DLS at the end of the trajectory are visibly deflected by the external tip force. This corresponds to the CMEs shown in Fig.\ref{fig:stiffness_control_force}-(B) and the change of $\mathbf{\mu}^d_c$ shown in Fig.\ref{fig:stiffness_control_force}-(C). As the robot moved forward, the TSGP slightly reduces the compliance in the vertical direction, which can compensate for the shape deforamtion and trajectory deviation induced by the external force. By contrast, both RR and DLS methods lack the capability to follow the desired trajectory with the external loads.

\section{Conclusion} \label{conclusion}

In this paper, we present a redundancy resolution framework for CTR based on an efficient method for calculating the gradient of CTR manipulability. Task-specific performance indices based on velocity/compliance manipulability is proposed for trajectory tracking in different operation scenarios. The proposed derivative propagation method reduces the computational time for the Hessian by 68\% compared to the finite difference method. 
Simulation studies were conducted in three specific scenarios corresponding to avoiding singularity, avoiding obstacles, and overcoming external force. The proposed redundancy resolution scheme consistently outperformed the standard resolved-rates and the damped least square method for trajectory tracking, demonstrating potential in facilitating teleoperation as well as task planning. 
Future work includes: 1) implementing the proposed method for hardware experiments and evaluating the real-time performance, 2) developing a high-level planner for autonomous task execution.

\ifCLASSOPTIONcaptionsoff
  \newpage
\fi

\bibliographystyle{ieeetr}
\bibliography{references}

\end{document}